\documentclass[11pt,a4paper]{article}
\usepackage{times,latexsym}
\usepackage{url}
\usepackage[normalem]{ulem}
\usepackage[T1]{fontenc}
\usepackage{booktabs}
\usepackage{tabularx}
\usepackage{relsize}
\usepackage{graphicx}
\usepackage{enumitem}
\usepackage{amsmath}
\usepackage{amssymb}
\usepackage{todonotes}

\usepackage[acceptedWithA]{tacl2018v2}

\usepackage{xspace,mfirstuc,tabulary}

\newif\iftaclinstructions
\taclinstructionsfalse 
\iftaclinstructions
\renewcommand{\confidential}{}
\renewcommand{\anonsubtext}{(No author info supplied here, for consistency with
TACL-submission anonymization requirements)}
\newcommand{\instr}
\fi

\iftaclpubformat 

\else

\fi

\setlist{itemsep=0pt,partopsep=0pt, parsep=0pt}

\title{Critical Thinking for Language Models}

\author{
 Gregor Betz\textsuperscript{\dag} \and 
 Christian Voigt\textsuperscript{\dag} \and
 Kyle Richardson\textsuperscript{\ddag}\\
 \textsuperscript{\dag} Karlsruhe Institute of Technology, Karlsruhe, Germany \\
\{gregor.betz, christian.voigt\}@kit.edu \\
 \textsuperscript{\ddag}Allen Institute for AI, Seattle, WA, USA \\
\{kyler\}@allenai.org\\ 
  
}

\date{}

\begin{document}
\maketitle

\begin{abstract}
This paper takes a first step towards a critical thinking curriculum for neural auto-regressive language models. We introduce a synthetic corpus of deductively valid arguments, and generate artificial argumentative texts to train and evaluate {\smaller GPT-2}. Significant transfer learning effects can be observed: Training a model on three simple core schemes allows it to accurately complete conclusions of different, and more complex types of arguments, too. The language models generalize the core argument schemes in a correct way. Moreover, we obtain consistent and promising results for NLU benchmarks. In particular, pre-training on the argument schemes raises zero-shot accuracy on the GLUE diagnostics by up to 15 percentage points. The findings suggest that intermediary pre-training on texts that exemplify basic reasoning abilities (such as typically covered in critical thinking textbooks) might help language models to acquire a broad range of reasoning skills. The synthetic argumentative texts presented in this paper are a promising starting point for building such a ``critical thinking curriculum for language models.''

\end{abstract}

\section{Introduction}

Pre-trained autoregressive language models (LM) such as {\small GPT-2} and {\small GPT-3} achieve, remarkably, competitive results in a variety of language modeling benchmarks without task-specific fine-tuning \citep{Radford2019,brown2020language}. Yet, it is also widely acknowledged that these models struggle with reasoning tasks, such as natural language inference (NLI) or textual entailment \cite{Askell:2020fb}. Actually, that doesn't come as a surprise, given the tendency of humans to commit errors in reasoning \cite{Kahneman2011,SunsteinHastie2015}, their limited critical thinking skills \cite{Paglieri2017}, the resulting omnipresence of fallacies and biases in texts and the frequently low argumentative quality of online debates \cite{hansson2004fallacies,guiacsu2018logical,Cheng:2017ud}. Neural language models are known to pick up and reproduce \emph{normative} biases (e.g., regarding gender or race) present in the dataset they are trained on \cite{Gilburt:2019tn}, as well as other \emph{annotation artifacts} \citep{gururangan2018annotation}; no wonder this happens with \emph{argumentative} biases and reasoning flaws, too \citep{kassner2020negated,talmor2020olmpics}. This diagnosis suggests that there is an obvious remedy for LMs' poor reasoning capability: make sure that the training corpus contains a sufficient amount of exemplary episodes of sound reasoning. 

In this paper, we take a first step towards the creation of a ``critical thinking curriculum'' for neural language models. Critical thinking can be loosely defined as ``reasonable reflective thinking that is focused on deciding what to believe or do.'' \citep{norris1989critical} Generally speaking, our study exploits an analogy between teaching critical thinking to students and training language models so as to improve their reasoning skill. More specifically, we build on three key assumptions that are typically made in critical thinking courses and textbooks: First, there exist fundamental reasoning skills that are required for, or highly conducive to, a large variety of more specific and advanced critical thinking skills \citep[e.g.,][p.~7]{Fisher2001CT}. Second, drawing deductive inferences is one such basic ability \citep[e.g.,][pp.~7--8]{Fisher2001CT}. Third, reasoning skills are not (just) acquired by learning a theory of correct reasoning, but by studying lots of examples and doing ``lots of good-quality exercises'' \citep{Lau2020CTW}, typically moving from simple to more difficult problems \citep[e.g.,][]{BowllKemp2014}.

These insights from teaching critical thinking translate, with respect to our study, as follows. First of all, we design and build `lots of good-quality exercises': a synthetic corpus of deductively valid arguments which instantiate a variety of (syllogistic) argument schemes, and which are rendered as text paragraphs (Section~\ref{sec:aac}). Next, we use our synthetic argument text corpus to train and to evaluate {\small GPT-2} (Section~\ref{sec:experiments}). The training, which maximizes a causal language modeling objective, can be conceived of as a generic, intermediary pre-training in the spirit of STILTS \citep{phang2018sentence}. 

Evaluating the models' ability to correctly complete conclusions of arguments, we observe strong transfer learning effects/generalization (Section~\ref{sec:results}): Just training the models on a few central core schemes (generalized modus ponens, contraposition and chain rule) allows them to accurately complete conclusions of different types of arguments, too (e.g., complex argumentative forms that involve dilemma and de Morgan). The language models appear to connect and generalize the core argument schemes in a correct way. In addition, the models are equally able to apply learned argument patterns beyond the training corpus' domain. Tests with a simple manually authored argument produce evidence that generic language modeling skill facilitates the successful generalization of learned argument patterns.

Moreover, we test the trained models on different reasoning benchmarks. Because we are particularly interested in transfer learning effects, we do so in a zero-shot set-up (i.e., evaluating our argumentation models on entirely unrelated NLU tasks, which follows recent work by \citet{mitra2019additional,shwartz2020unsupervised,ma2020knowledge}). We obtain consistent and promising results for the GLUE diagnostics \citep{wang2018glue} and SNLI \citep{snli:emnlp2015} benchmarks (Section~\ref{sec:results}), finding that training on core schemes clearly improves NLU skill. However, training on the argument corpus doesn't affect the performance with regard to the semantically more demanding Argument Reasoning Comprehension task \citep{DBLP:conf/naacl/HabernalWGS18a} or the critical thinking assessment compiled in LogiQA \citep{DBLP:conf/ijcai/LiuCLHWZ20}. 

All these transfer learning effects observed strengthen the analogy between teaching critical thinking and training language models: A variety of reasoning skills are improved by generic, intermediary pre-training on high-quality texts that exemplify a basic reasoning skill, namely simple deductive argumentation. Obviously, drawing correct inferences is just one of the elementary skills typically covered in critical thinking courses \citep{Fisher2001CT}. Critical thinking involves more than deduction. And it would hence, by analogy, be unreasonable to expect that intermediary pre-training on the synthetic argument corpus suffices to turn language models into accomplished reasoners. However, we have shown that argumentative texts (with valid syllogistic arguments) are certainly a good starting point when building a more comprehensive dataset for initial or intermediary pre-training that might help language models to acquire a broad range of reasoning skills. Or, to put it differently, the synthetic argumentative texts might belong to the core of a ``critical thinking curriculum for language models.'' In the final section, we advance some ideas for complementing the artificial argument corpus so as to further improve the performance of LMs with regard to different reasoning benchmarks.


\section{Related Work}

To our knowledge, this paper is, together with \citet{gontier2020measuring}, among the first to show that autoregressive language models like {\small GPT-2} can learn to reason by training on a \emph{text corpus} of correct natural language arguments. By contrast, previous work in this field, described below, has typically modeled natural language reasoning problems as classification tasks and trained neural systems to accomplish them. For example, \citet{schick2020exploiting, schick2020its}, using pattern verbalizations, construct structured training data that is suitable for training a \emph{masked} language model with classification head, and thusly achieve remarkable NLU performance. This paper explores the opposite route: We start with highly structured (synthetic) data, render it as unstructured, plain text and train a \emph{uni-directional} language model on the synthetic text corpus.


Over and above the methodological novelty of our approach, we discuss, in the following, related reasoning benchmarks and explain what sets our synthetic argument corpus apart from this work.   

\paragraph{Rule reasoning in natural language} Various datasets have been developed for (deductive) rule reasoning in natural language. In these tasks, one or multiple rules, i.e. (generalized) conditionals, must be applied to a fact base in order to deductively infer a conclusion. Facts and conclusions are represented by atomic statements. Rule application closely resembles the conclusion completion task for \emph{generalized modus ponens} and \emph{generalized modus tollens} schemes described below. However, we go beyond previous work in investigating the ability of language models to infer conclusions that have a more complex logico-semantic structure (e.g., existential or universal statements).

The question answering bAbI dataset \cite{Weston:2016dy} contains a task which involves applying very specific rules of the form “Xs are afraid of Ys” to an instance (for example: ``Mice are afraid of cats. Jerry is a mouse. What is Jerry afraid of? \emph{A:cats}''). Equally simple, one-step rule applications are tested in \citet{Richardson:2020ym}, and also contained in the QuaRTz dataset \cite{Tafjord:2019xa}.

ROPES \cite{Lin:2019tw} is a reading comprehension task that involves applying background knowledge to a given situation (both being presented as paragraph long text). Correct answers can be inferred by one-step rule application; part of the challenge is to identify the relevant rule and fact in the text.     

RuleTaker, arguably the most general system for natural rule reasoning in natural language so far, is a transformer model that has been fine-tuned to predict whether a conclusion can be inferred from a set of rules and facts, not all of which are necessarily required to draw the conclusion \citep{Clark2020_TransSoftReas}. Moreover, inferring the conclusion from the premise set might involve multiple inference steps. The authors show that the transformer model can be trained to perform this task nearly flawlessly and, moreover, to `explain' its inferences by identifying relevant premises. They also observe substantial transfer learning effects.

PRover extends RuleTaker by a component for proof generation \citep{saha2020prover}. Technically, the QA head of the RoBERTa language model \citep{liu2019roberta} is complemented by two additional neural classifiers (for nodes and edges) that are used to to construct proof chains. \citet{saha2020prover} show that PRover can construct valid proofs and outperforms RuleTaker in terms answer accuracy in a zero-shot setting.

Training on synthetic knowledge-graph data (such as "Paris CapitalOf France" and "France HasCapital Paris") \emph{from scratch}, \citet{kassner2020pretrained} find that BERT is able to correctly infer novel facts. This confirms that language models can, in principle, learn basic conceptual rules, which, e.g., express that a relation is symmetric or that two terms are equivalent.

\paragraph{Benchmarks for enthymematic reasoning} An `enthymeme' is an argument whose premises are not explicitly stated, e.g.: ``Jerry is a mouse. Therefore, Jerry is afraid of cats.'' The three tasks described below involve such reasoning with implicit assumptions, whereas our synthetic argument corpus doesn't: all premises are transparent and explicitly given. 

Commensense Transformers (COMET) are autoregressive language models for generating commonsense knowledge graphs \citep{Bosselut2019COMETCT}. Being trained on seed data, the models are able to meaningfully relate subject phrases to object phrases in terms of multiple binary relations (by doing the type of completion tasks we introduce in Section~\ref{sec:experiments}), and can thereby both reproduce and extend a given knowledge graph. In particular, this includes generating statements about causal relationships, which can be construed as enthymematic reasoning with commonsense background assumptions. For example, given the input "PersonX is re-elected. As a result, PersonX wants" the model generates as completions: "to get a raise", "to go to office", "to go home", "to make a speech", "to celebrate" -- all of which are plausible fill-ins. The implicit commonsense premises that underlie this (entyhmematic) inference are principles such as "If someone has been re-elected, then they want to celebrate."

The Argument Reasoning Comprehension (ARC) dataset \cite{DBLP:conf/naacl/HabernalWGS18a} comprises simple informal arguments. Each argument contains two premises: whereas the first premise is explicitly stated, there are two alternative formulations of the second premise. The task consists in identifying which of these two alternative formulations is actually assumed in the argument. For example: ``Miss America gives honors and education scholarships. And since [\emph{scholarships would give women a chance to study} $|$ \emph{scholarships would take women from the home}], Miss America is good for women.'' ARC therefore assesses the ability to make implicit premises explicit. An adversarial ARC dataset that eliminates clues in the original benchmark is also available in \citet{niven-kao-2019-probing}. 
 
CLUTRR is a task generator for relational reasoning on kinship graphs \cite{Sinha:2019va}. CLUTTR takes a set of (conceptual) rules about family relations as given and constructs set-theoretic possible worlds (represented as graphs) which instantiate these rules. In such a possible (kinship) world, a target fact and a set of base facts are identified such that the base facts together with the rules deductively entail the target fact. The task consists in inferring the target fact from the base facts alone -- the conceptual rules remain implicit. For example: ``Kristin and her son Justin went to visit her mother Carol on a nice Sunday afternoon. They went out for a movie together and had a good time. Q: How is Carol related to Justin? A: Carol is the grandmother of Justin.'' So, CLUTRR assesses entyhmematic deductive reasoning with implicit conceptual rules. \citet{gontier2020measuring} have trained a generative Transformer language model on a synthetic text corpus (with each argumentative text containing a story, a proof chain and a conclusion from CLUTTR) and show that the language model does not only learn to draw the correct conclusion (given an argument with implicit commonsense premises), but also seems to acquire the ability to generate valid proof chains.

\paragraph{Critical thinking tasks}

LogiQA \cite{DBLP:conf/ijcai/LiuCLHWZ20} is a collection of publicly available critical thinking questions, used by the National Civil Servants Examination of China to assess candidates’ critical thinking and problem solving skills. LogiQA covers tasks of various types: different kinds of natural language inference problems as well as the identification of implicit premises or (practical) instrumental reasoning. Its scope is much broader than our highly specific and carefully designed argument corpus.  The LogiQA tasks are shown to be hard for current AI systems, of which a fine-tuned transformer model performs best with an accuracy score of 35\% -- 50 percentage points below human performance.

\section{An Artificial Argument Corpus}\label{sec:aac}

This section describes the construction of a synthetic corpus of natural language arguments used for training and evaluating {\small GPT-2}.\footnote{The corpus as well as the source code used to generate it will be released at \url{https://github.com/debatelab/aacorpus}.}

\begin{figure*}[htb]
    \centering
    \includegraphics[width=\textwidth]{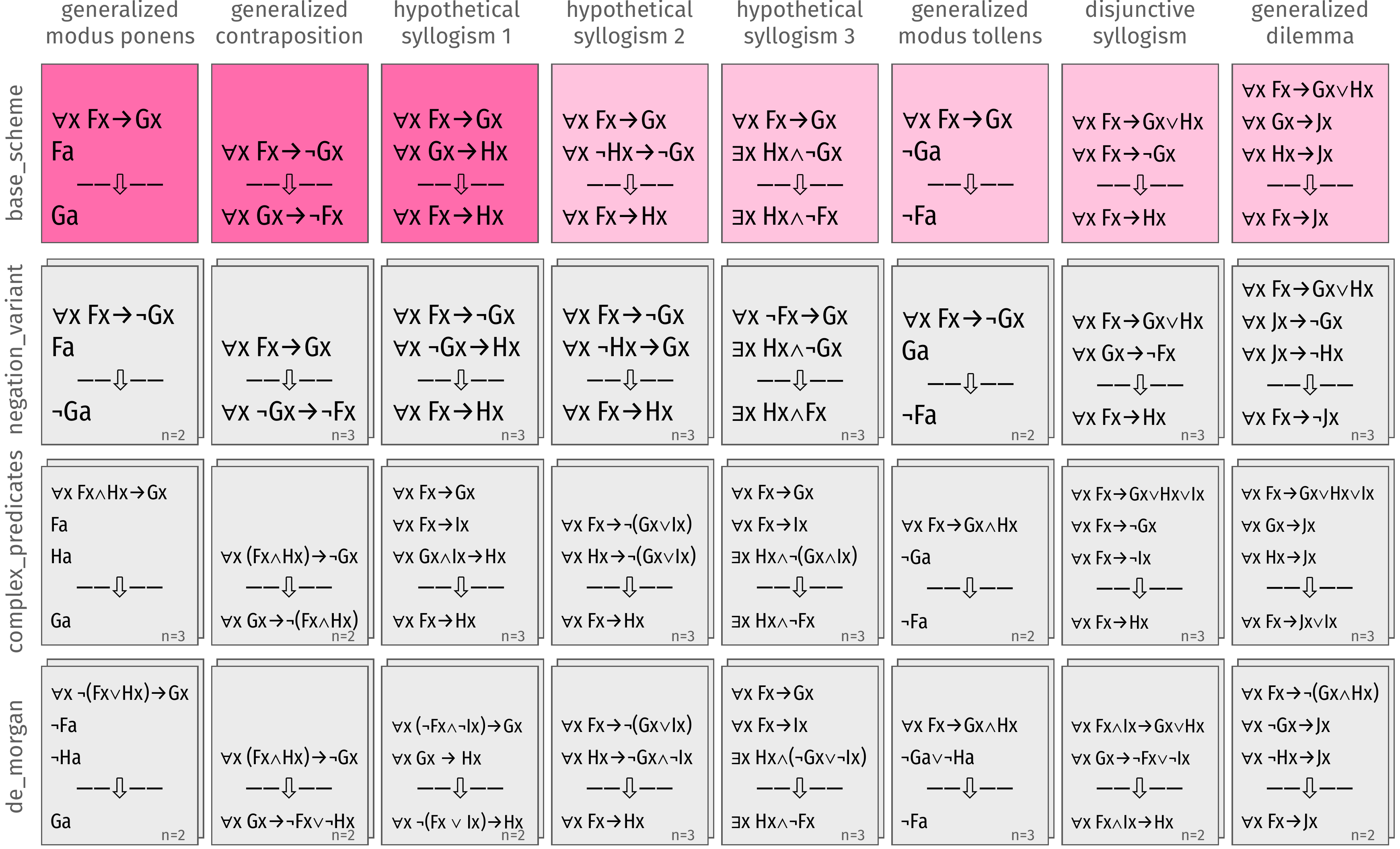}
    \caption{Syllogistic argument schemes used to create an artificial argument corpus.}
    \label{fig:arg_schemes}
\end{figure*}

The corpus is built around eight simple, deductively valid syllogistic argument schemes (top row in Figure~\ref{fig:arg_schemes}). These \emph{base schemes} have been chosen because of their logical simplicity as well as their relevance in critical thinking and argument analysis \cite{Feldman2014,BowllKemp2014,BrunBetzRaU2016}. Each of these eight base schemes is manually varied in specific ways to create further valid variants.

\emph{Negation variants} of base schemes (second row in Figure~\ref{fig:arg_schemes}) are created by substituting a sub-formula with its negation and/or by applying \emph{duplex negatio affirmat}.

\emph{Complex predicates} variants (third row in Figure~\ref{fig:arg_schemes}) build on base schemes or their respective negation variants and are obtained by substituting atomic predicates with compound disjunctive or conjunctive ones.

\emph{De Morgan} variants of base schemes (fourth row in Figure~\ref{fig:arg_schemes}) are finally derived by applying de~Morgan's law to the respective variants created before. 

With 2-3 different versions for each of these variations of a base scheme (parameter "n" in Figure~\ref{fig:arg_schemes}), we obtain, all in all, 71 distinct handcrafted argument schemes. Obviously, some of these schemes can be derived from others. For example, generalized modus ponens and generalized contraposition (base schemes) entail a \emph{negation variant} of generalized modus tollens. Likewise, generalized contraposition and hypothetical syllogism 1 entail a (\emph{negation variant} of) hypothetical syllogism 2.

In view of their simplicity and prominence in natural language argumentation, three of the eight \emph{base schemes} are marked as \emph{core schemes}: generalized modus ponens, generalized contraposition, hypothetical syllogism 1.

\begin{figure*}[htbp]
    \centering
    \includegraphics[width=\textwidth]{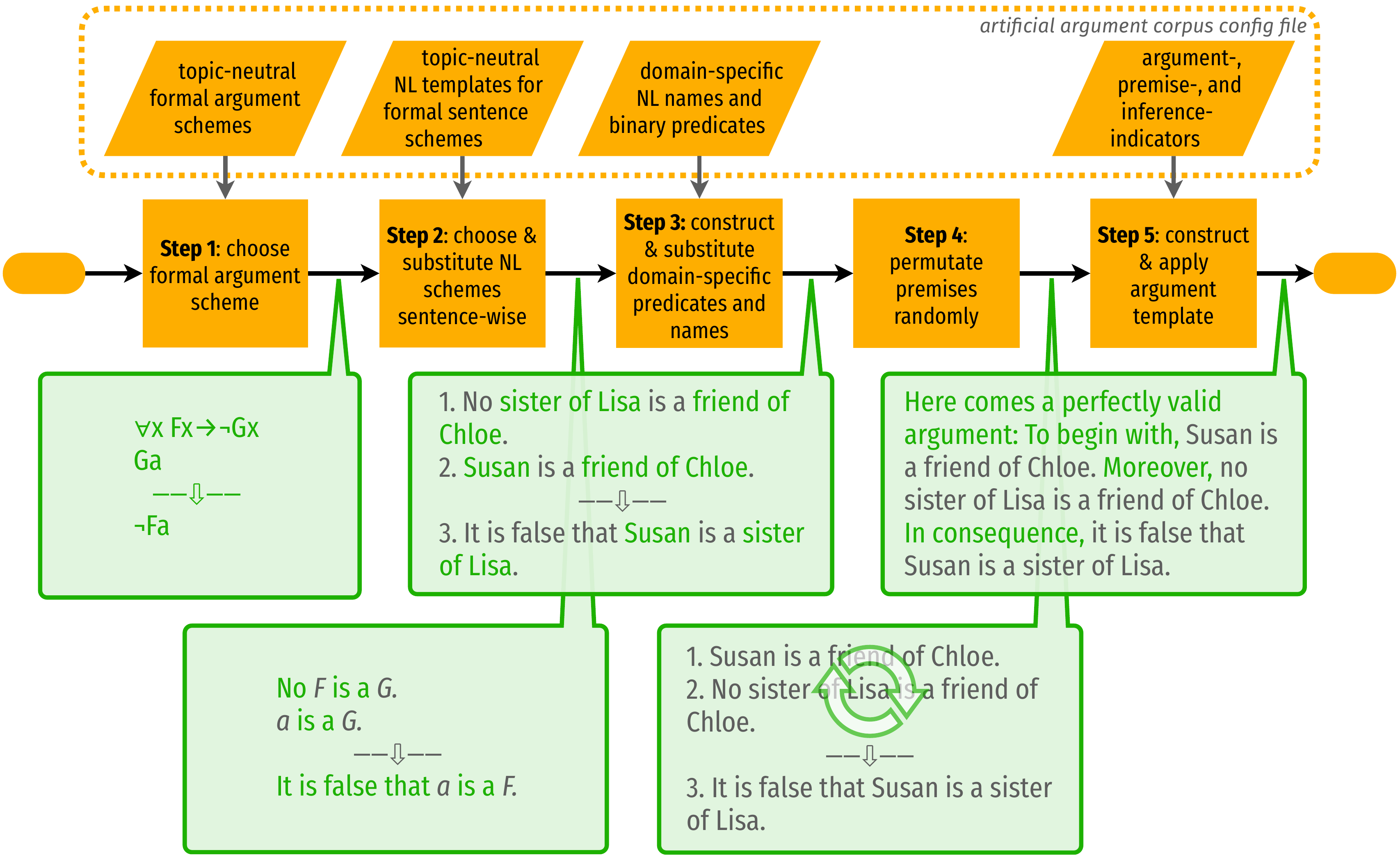}
    \caption{Pipeline for creating natural language instances of argument schemes with multiple templating.}
    \label{fig:pipeline}
\end{figure*}

Natural language instances of the argument schemes can be created by means of a first-order-logic domain (with names and predicates) and natural language templates for the formal schemes. In order to obtain a large variety of realistic natural language arguments, we have devised
\begin{itemize}
    \item a multi-stage templating process with
    \item alternative templates at each stage and
    \item multiple domains.
\end{itemize}
As shown in Figure~\ref{fig:pipeline}, this process can be split into five consecutive steps. 

In \emph{step 1}, the argument scheme, which serves as formal template for the natural language argument, is chosen.

In \emph{step 2}, each sentence in the formal scheme (premises and conclusion) is individually replaced by a natural language pattern in accordance with a randomly chosen template. For example, the formula ``$\forall x Fx \rightarrow Gx$'' might be replaced by any of the following natural language sentence schemes:
\begin{itemize}
    \item ``Every F is a G.''
    \item ``Whoever is a F is also a G.''
    \item ``Being a G is necessary for being a F.''
    \item ``If someone is a F, then they are a G.''*
\end{itemize}
Some of these patterns are not used for training, but are reserved for generating an out-of-domain test dataset (e.g., the template marked with an asterisk in the above list).  

In \emph{step 3}, the entity- and property-placeholders in the resulting argument scheme are replaced argument-wise with names and predicates from a domain. We hence obtain an instance of the formal argument scheme as premise-conclusion list. Each domain provides hundreds of entity-names, which can be paired with different binary predicates to create thousands of different unary predicates. The following example predicates illustrate the domains used in this study:
\begin{itemize}
    \item \emph{Female Relatives:} sister of Anna, granddaughter of Elsa, cousin of Sarah, \ldots
    \item \emph{Male Relatives:} grandson of Ryan, nephew of Jim, cousin of Lee, \ldots
    \item \emph{Football Fans:} supporter of Real Madrid CF, ex-fan of Sevilla FC, member of SSC Napoli, \ldots
    \item \emph{Personal Care:} regular consumer of Dove shampoo, infrequent user of L'Oreal shampoo, loyal buyer of Redken shampoo, \ldots
    \item \emph{Chemical Ingredients:} ingredient of Maypole Soap, ingredient of OASIS CREAM, ingredient of BB concealer, \ldots
    \item \emph{Dinosaurs*:} contemporary of Megalosaurus, predator of Iguanodon, ancestor of Allosaurus, \ldots
    \item \emph{Philosophers*:} teacher of Aeschines of Neapolis, pupil of Cratylus, reader of Democritus, \ldots
\end{itemize}
Domains marked with an asterisk are used for testing only, and not for training (see below and Section~\ref{subsec:testing}). 

In \emph{step 4}, the premises of the natural language argument are randomly re-ordered.

In \emph{step 5}, the premise-conclusion list is packed into a text paragraph by adding an argument intro, framing the premises, and adding an inference indicator. Again, multiple templates are available for doing so, which yields a large variety of textual renderings of an argument.  

Following this pipeline, we generate natural language instances of each formal argument scheme, thus creating:
\begin{enumerate}
    \item a training set of argumentative texts, based on the default domains and templates ({\small TRAIN});
    \item an evaluation set of argumentative texts, based on the default domains and templates, which are used for development ({\small DEV});
    \item a test set of argumentative texts, based on the default domains and templates and used for final tests ({\small TEST\_OUT-OF-SAMPLE});
    \item a test set of argumentative texts, based on the domains and templates reserved for testing ({\small TEST\_OUT-OF-DOMAIN}).
\end{enumerate}

This represents the artificial argument text corpus we use to train and evaluate {\small GPT-2}.

\section{Experiments with {\smaller GPT-2}} \label{sec:experiments}

We train and evaluate three compact versions of {\small GPT-2} with 117M, 345M and 762M parameters respectively using the implementation from \citet{wolf2019huggingface}. We note that all of these models fall short of the full-scale model with 1542M parameters.\footnote{The fine-tuned models will be released through \url{https://huggingface.co/models}.} 

\subsection{Training}


From the training items in the Artificial Argument Corpus ({\small TRAIN}) we sample three types of differently-sized training sets as follows (see also the color pattern in Figure~\ref{fig:arg_schemes}): 
\begin{itemize}
  \item {\small TRAIN01}: all training items which are instances of a \emph{core scheme}, i.e. generalized modus ponens, generalized contraposition, hypothetical syllogism 1 (N=4.5K, 9K, 18K, 36K)
  \item {\small TRAIN02}: all training items which are instances of a \emph{base scheme} (N=4.5K, 9K, 18K, 36K)
  \item {\small TRAIN03}: all training items in the corpus (N=4.5K, 9K, 18K, 36K)
\end{itemize} 

In an attempt to avoid over-fitting, we blend the training arguments with snippets from Reuters news stories \cite{Lewis:2004ux} and the standardized Project Gutenberg Corpus \cite{DBLP:journals/corr/abs-1812-08092}, trying a mixing ratio of 1:1 and thus doubling training size to N=9K, 18K, 36K, 72K. (We find that fine-tuning on the accordingly enhanced argument corpus still increases the model's perplexity on the Wiki103 dataset by a factor of 1.5 (see Appendix~\ref{appendix:training_size}), which suggests to mix a higher proportion of common texts into the training data in future work.) The three different versions of {\small GPT-2} are fine-tuned (causal language modeling objective, using default training scripts by \citet{wolf2019huggingface}) on each of the 12 enhanced training sets (hyper-parameters are detailed in Appendix~\ref{appendix:hyperparameters}). This gives us 36 fine-tuned model versions plus the three {\small BASE} models to evaluate. Unless explicitly stated otherwise, we report results of 762M parameter model trained on 72K items.  

\subsection{Testing}\label{subsec:testing}

\paragraph{Conclusion Completion on Artificial Argument Corpus} To test whether language models can reason correctly, we assess their ability to accurately complete conclusions of arguments in the artificial argument corpus. Here, we make use of the fact that, by construction, the conclusion of every argument in the corpus ends with a predicate (a property-term such as ``sister of Chloe'' or ``supporter of Tottenham Hotspurs''), which is potentially preceded by a negator. First of all, as shown in Table~\ref{table:cctasks}, we test whether the model is able to correctly fill in the final predicate (task \emph{split}). The second, more difficult task consists in completing the final predicate plus, if present, the preceding negator (task \emph{extended}). With a third, adverserial task we check how frequently the model wrongly adjoins the complement of the correct completion of the \emph{extended} task (task \emph{inverted}). Consider, for example, the following argument: 

\begin{quote}
It is not always easy to see who is related to whom -- and in which ways. The following argument pertains to this question: First premise: Every workmate of Brad is a classmate of James. Second premise: Every classmate of James is not a classmate of Theodore. So, necessarily, everyone who is a workmate of Brad is [\underline{not a}]$_E$ [\underline{classmate of Theodore.}]$_S$''
\end{quote}
In the \emph{split} task, we prompt the model with the argument, dropping []$_S$, and check whether it generates ``classmate of Theodore''. In the \emph{extended} task, we prompt the model with the argument, dropping []$_E$[]$_S$, and check whether it generates ``not a classmate of Theodore''. Finally, in the \emph{inverted} task, we prompt the model as before and check whether it generates ``a classmate of Theodore''. 

\begin{table}[htbp]
  \label{table:ruletakerexample}
  \begin{tabular}{@{}p{0.23\linewidth}@{\rule[0mm]{0.05\linewidth}{0cm}}p{0.47\linewidth}@{\rule[0mm]{0.03\linewidth}{0cm}}p{0.17\linewidth}}
    \toprule
    Task & Conclusion with \fbox{cloze-style prompt} \raggedright & Comple\-tion \\ 
    \midrule
    \textit{split}  & \fbox{Every \textit{F} is a} \textit{G}        & \textit{G} \\
                    & \fbox{Some \textit{F} is not a} \textit{G}     & \textit{G} \\
                    & \fbox{\textit{a} is a \textit{F} or not a} \textit{G}       & \textit{G} \\
    \midrule
    \textit{extended}   & \fbox{Every \textit{F} is} a \textit{G}        & a \textit{G} \\
                        & \fbox{Some \textit{F} is} not a \textit{G}     & not a \textit{G} \\
                        & \fbox{\textit{a} is a \textit{F} or} not a \textit{G}       & not a \textit{G} \\
    \midrule
    \textit{inverted}   & \fbox{Every \textit{F} is} a \textit{G}        & \underline{not} a \textit{G} \\
                        & \fbox{Some \textit{F} is} not a \textit{G}     & \sout{not} a \textit{G} \\
                        & \fbox{\textit{a} is a \textit{F} or} not a \textit{G}       & \sout{not} a \textit{G} \\
    \bottomrule
  \end{tabular}
  \caption{Three conclusion completion tasks}\label{table:cctasks}
\end{table}

Clearly, the higher the accuracy in the \emph{split} and \emph{extended} tasks, and the lower the accuracy in the \emph{inverted} task, the stronger the model's reasoning performance.

Based on the artificial argument corpus (see Section~\ref{sec:aac}), we generate and distinguish three different test datasets, each of which comprises the three tasks described above, as follows:

\begin{itemize}
    \item \emph{out of sample}: contains items from {\small TEST\_OUT-OF-SAMPLE}, which share domain and natural language templates with the training data;  
    \item \emph{paraphrased}: a sample of 100 items, randomly drawn from {\small TEST\_OUT-OF-SAMPLE}, which have been manually reformulated so as to alter the premises' grammatical structure imposed by the natural language templates;
    \item \emph{out of domain}: contains items from {\small TEST\_OUT-OF-DOMAIN}, which belong to different domains instantiate grammatical patterns other than the training data. 
\end{itemize}

Technically, conclusion completions, in all tasks and tests, are generated by the language model with top-p nucleus sampling ($p=0.9$). 

\paragraph{Classification for NLU Benchmarks}

To investigate transfer learning effects, we evaluate the trained models on standard NLU benchmarks, such as GLUE AX and SNLI. These benchmark tasks are classification problems. In the following, we describe how we use the generative language models to perform such classification. 

Using simple templates, we translate each benchmark entry into alternative prompts (e.g., context and question) and/or alternative completions (e.g., answers). Consider for example a GLUE-style problem given by two sentences ``The girl is eating a pizza.'' and ``The girl is eating food'' and the question whether one entails, contradicts, or is independent of the other. We can construct three prompts, corresponding to the three possible answers (entail / contradict / independent):
\begin{quotation}
\noindent\emph{Prompt1}: The girl is eating a pizza. Therefore, \\
\emph{Prompt2}: The girl is eating a pizza. This rules out that \\ 
\emph{Prompt3}: The girl is eating a pizza. This neither entails nor rules out that \\ 
\emph{Completion}: the girl is eating food.
\end{quotation}
In this case, the correct match is obviously \emph{Prompt1}--\emph{Completion}. The ability of a  language model to discern that ``The girl is eating pizza'' entails (and does not contradict) ``The girl is eating food'' will be reflected in a comparatively low conditional perplexity of \emph{Completion} given \emph{Prompt1} and a correspondingly high conditional perplexity of \emph{Completion} given \emph{Prompt2} or \emph{Prompt3}.

Let us describe this procedure in more general terms and consider a textual classification problem with categories $k=1\ldots N$. To classify a given input $X$, one constructs $n$ alternative prompts $p_1,\ldots p_n$ and $m$ alternative completions $c_1,\ldots,c_m$ ($N=m\cdot n$), such that each pair $(p_i,c_j)$ corresponds to a class $k$ of the classification problem, i.e., 

$$L: (p_i,c_j) \mapsto \{1\ldots N\}.$$

In the above pizza example, we have $N=n=3$ and $m=1$. Moreover, let $\mathrm{PP}_\mathbb{L}(c|p)$ refer to the conditional perplexity of the completion $c$ given prompt $p$ according to the language model $\mathbb{L}$. Rather than directly using this conditional perplexity as a prediction score \citep[as for instance in][]{shwartz2020unsupervised}, which doesn't account for varying `prima facie' or `prior' perplexities of alternative completions, we consider the degree to which prompting the model $\mathbb{L}$ with $p$ changes the the perplexity of $c$, i.e. 

$$\mathrm{relPP}_\mathbb{L}(c,p) := \frac{\mathrm{PP}_\mathbb{L}(c|p)}{\mathrm{PP}_\mathbb{L}(c)}.$$

In analogy to Bayesian confirmation theory, this might be termed a (perplexity-based) \emph{relevance measure}, as opposed to a measure of absolute confirmation \citep[cf.][pp.~346-48]{carnap:1950}.
We now use relevance perplexity as a score function to predict the category of $X$: 

$$\mathrm{category}(X) = L\big(\underset{(p_i,c_j)}{\mathrm{argmin}}(\mathrm{relPP}(c_j,p_i))\big).$$

\section{Results}\label{sec:results}

\paragraph{Conclusion Completion on Artificial Argument Corpus}

\begin{figure*}[htbp]
    \centering
    \includegraphics[width=\textwidth]{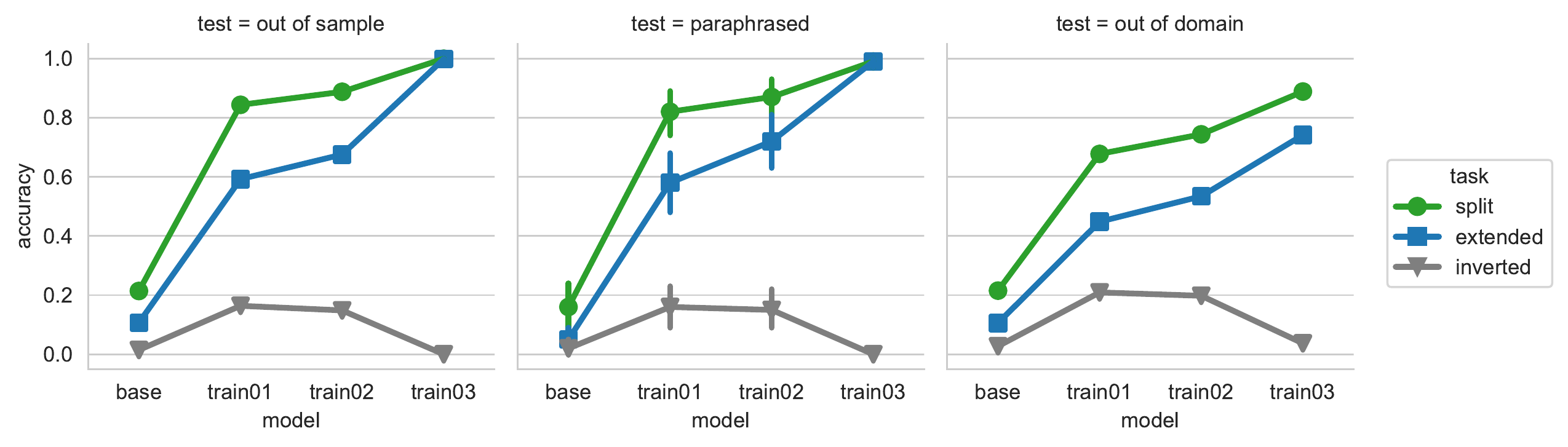}
    \caption{Accuracy of four model versions in three conclusion completion tasks and on different test datasets (out of sample, paraphrased, out of domain).}
    \label{fig:eval_row}
\end{figure*}

Does the (fine-tuned) {\small GPT-2} model correctly complete conclusions of natural language arguments? Figure~\ref{fig:eval_row} displays the evaluation results in an aggregated way. Each subplot visualizes the accuracy of the models in the three completion tasks for a different test dataset (see Section~\ref{subsec:testing}), comparing the {\small BASE} model (points at the very left) with the fine-tuned models trained on {\small TRAIN01}, {\small TRAIN02}, and {\small TRAIN03} (in this order from left to right). The task-specific accuracy values are distinguished by line color. 
  
We may observe, first of all, that training on the argument corpus effectively improves conclusion-completion-skill. In all three test datasets, the accuracy in the \emph{split} and \emph{extended} tasks increases as the models are trained on more and more argument schemes, far exceeding the base model's performance. Once the model has seen all schemes ({\small TRAIN03}), accuracy levels reach 100\% for in-domain and 70\%-90\% for out-of-domain tests. However, the {\small TRAIN01} and {\small TRAIN02} models do also generate more incorrect completions than the {\small BASE} model (\emph{inverted} task). But the frequency of such incorrect completions increases much less than the frequency of correct ones (the gap between blue and gray curve widens), and it actually falls back to almost zero with the {\small TRAIN03} model.  Out-of-domain performance of the models (right-hand plot) is qualitatively similar and only slightly less strong than in-domain performance (left-hand and middle plot). The models trained on arguments from a given domain are able to effectively exercise the reasoning skill thus acquired in other domains, and have hence gained topic-neutral, universal reasoning ability.

\begin{table*}[htbp]
   \begin{center}\small
   \begin{tabularx}{\linewidth}{@{}p{2cm} c@{\rule{8mm}{0mm}}cccccc@{}}
      \toprule
      & {\smaller BASE} & \multicolumn{3}{c}{(a) schemes not in training data ({\smaller TR01--02})} & \multicolumn{3}{c}{(b) trained on schemes ({\smaller TR01--03})} \\
      \cmidrule(r){3-5} \cmidrule(l){6-8}
      Task    & & \emph{o-o-sample} & \emph{paraphr.} & \emph{o-o-domain} & \emph{o-o-sample} & \emph{paraphr.} & \emph{o-o-domain}\\
      \midrule
      
      \emph{split} & 21.4      &       85.4 &     82.0 &      69.4 &       99.9 &     99.2 &       89.0 \\
      \emph{extended}& 10.7    &       60.3 &     59.3 &      45.8 &      99.9 &     99.2 &       76.2 \\
      \emph{inverted} & 1.5 &       16.9 &     18.0 &      22.1 &      0.0 &      0.0 &       3.2 \\
      
      \bottomrule
   \end{tabularx}
   \end{center}
   \caption{Accuracy of models in three conclusion completion tasks and on different test datasets (out of sample, paraphrased, out of domain). Columns report, separately, the performance (a) on schemes the model has not been trained on, and (b) on schemes that are covered by the model's training data.}
   \label{table:outofscheme}
\end{table*}

The strong performance of {\small TRAIN01} models, averaged over all schemes, suggests that significant transfer learning occurs and that training on a few argument schemes positively affects performance on other schemes, too. To further investigate this issue, Table~\ref{table:outofscheme} contrasts (a) the models' accuracy on schemes they have not been trained on -- averaged over  {\small TRAIN01} and {\small TRAIN02} models -- with (b) their accuracy on schemes that are instantiated in their respective training corpus -- averaged over  {\small TRAIN01}, {\small TRAIN02}, and {\small TRAIN023} models. The upshot is that trained models perform way more strongly than the base model not only on argument schemes they've been trained, but also on those schemes they haven't seen yet. We take this to be a promising result as it strengthens the analogy between teaching critical thinking and training language models: generic intermediary pre-training on high-quality texts that exemplify a specific, basic reasoning skill -- namely, simple deductive argumentation -- improves other, more complex reasoning skills.

\begin{figure*}[htbp]
    \centering
    \includegraphics[width=\textwidth]{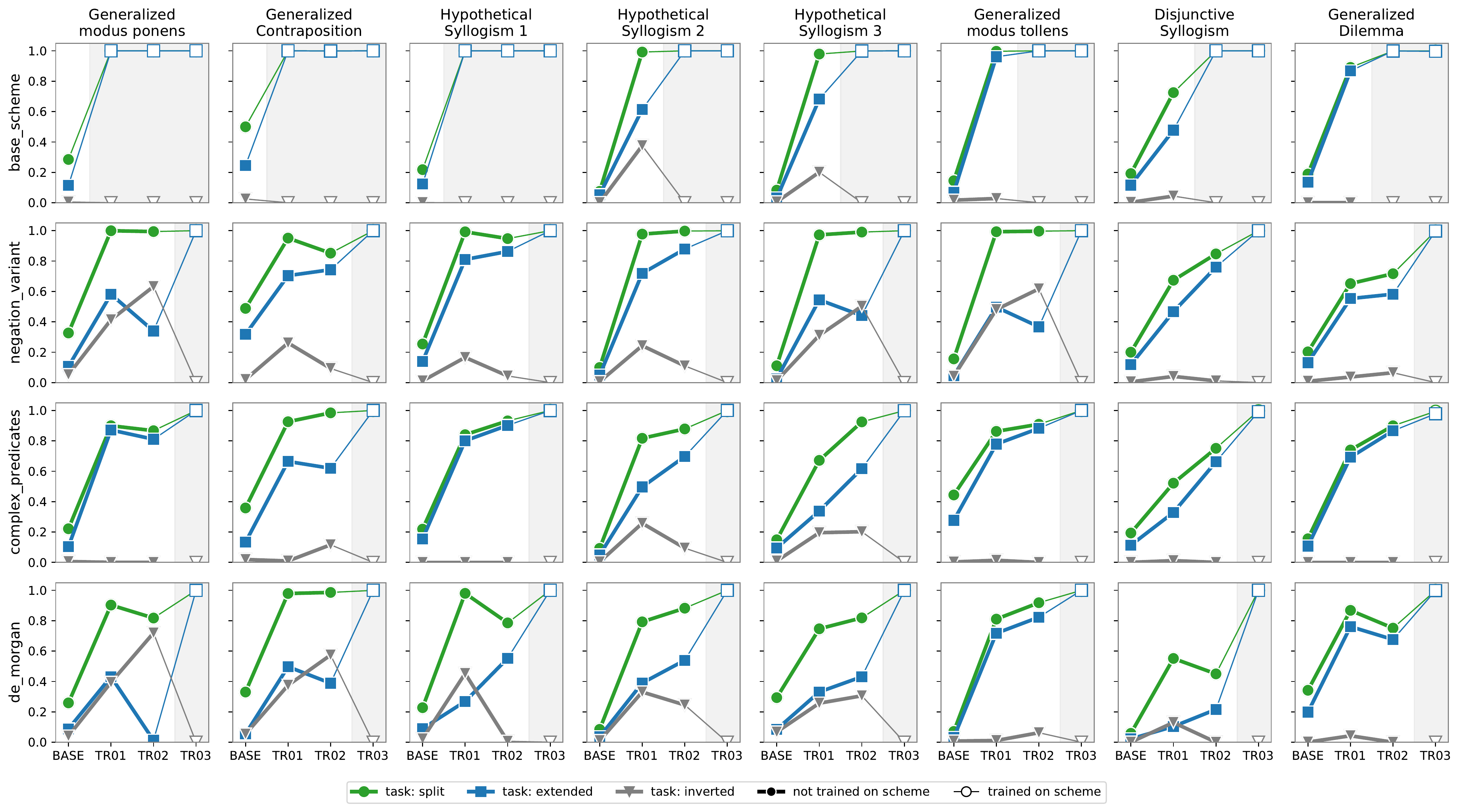}
    \caption{Accuracy of conclusion completions (three tasks) for instances of different argument schemes (see Figure~\ref{fig:arg_schemes}) and four model versions.}
    \label{fig:eval_matrix}
\end{figure*}

Figure~\ref{fig:eval_matrix} gives further insights by differentiating evaluation results according to argument type. Its subplots are arranged in a grid that mirrors the organisation of argument schemes in Figure~\ref{fig:arg_schemes}. Each subplot visualizes the ability of the models to correctly complete arguments of the corresponding scheme (given the out-of-sample test dataset). Accordingly, the left-hand plot in Figure~\ref{fig:eval_row} in effect averages all curves in Figure~\ref{fig:eval_matrix}. Reported accuracy values that fall within gray background areas are attained by models which have seen the corresponding scheme during training. Vice versa, thick lines on white background visualize model performance on unknown schemes. 
Figure~\ref{fig:eval_matrix} reveals, first of all, that even the {\small BASE} models (only pre-training, no fine-tuning) display a significant ability to correctly complete conclusions of some kinds of arguments. For example, {\small GPT-2}-762M achieves 50\% accuracy (\emph{split} task) in completing contrapositions, 30\% accuracy in completing generalized modus ponens, and still 20\% accuracy in completing disjunctive syllogism and dilemma arguments. These findings further corroborate the hypothesis that NLMs learn (basic) linguistic and reasoning skills ``on the fly'' by training on a large generic corpus \cite{Radford2019}. 

In addition, the matrix plot (Figure~\ref{fig:eval_matrix}) demonstrates that some types of arguments are much easier to master, given training on the core and possibly base schemes, than others. For instance, \emph{complex\_predicates} variants of generalized modus ponens or \emph{de\_morgan} variants of generalized modus tollens seem to be easily mastered by the {\small TRAIN01} model. In contrast, even the {\small TRAIN02} model, which has been fine-tuned on all eight base schemes, struggles with the \emph{negation\_variants} of generalized modus ponens (generating substantially more incorrect than correct completions). All in all, the picture that emerges is plausible: Generalization towards novel types of argument appears to be comparatively difficult whenever the new scheme involves negations (compare 2nd and 4th row in Figure~\ref{fig:eval_matrix} with 3rd row). This is consistent with the finding that some NLMs seemingly fail to understand simple negation \citep{kassner2020negated,talmor2020olmpics}.

The results reported so far suggest that reasoning skills acquired on (a subset of) the artificial argument corpus generalize rather well -- both to other domains and other types of arguments. We have further cross-checked these statistical findings by letting the models complete a conclusion of a simple manually authored argument:
\begin{quote}
\textbf{[Hermes]} Every philosopher is mortal. Hermes is not mortal. Therefore, Hermes \ldots
\end{quote}
This text differs syntactically and semantically from any argument possibly contained in the artificial argument corpus (where predicates have always the form ``is/being a Y of X,'' and no domain covers philosophers or mortality). Obviously, it follows that Hermes ``is not a philosopher.'' The argument instantiates \emph{generalized modus tollens}, which is not a core scheme in {\small TRAIN01}. Can {\small TRAIN01}-models nonetheless fill out the unfinished argument in a sensible way?  

\begin{table}[htbp]
   \begin{center}\small
   \begin{tabular}{@{}p{4.2cm}@{ }c@{ }rrr @{}}
      \toprule
      && \multicolumn{2}{c}{762M} & 117M \\
      \cmidrule(r){3-4}
      Completion    & & {\smaller TR01} & {\smaller BASE} & {\smaller TR01}\\
      \midrule
            \ldots is not a philosopher.                    & $\star$       &    100 &     2 &    2 \\
                     \ldots is immortal.                    & $=$           &      0 &    12 &    0 \\
                 \ldots is not a critic.                    & $\circ$       &      0 &     0 &    9 \\
                       \ldots is mortal.                    & $\dagger$     &      0 &     8 &    0 \\
                   \ldots is not mortal.                    & $=$           &      0 &     6 &    0 \\
                   \ldots is not Hermes.                    & $\dagger$     &      0 &     2 &    0 \\
                  \ldots does not exist.                    & $\circ$       &      0 &     2 &    0 \\
                      \ldots is not God.                    & $\circ$       &      0 &     2 &    0 \\
     \ldots is not a friend of Eckhardt.                    & $\circ$       &      0 &     0 &    1 \\
  \ldots is not an expert of BSI Arsenal FC.                & $\circ$       &      0 &     0 &    1 \\
     \ldots is not a friend of Atalanta.                    & $\circ$       &      0 &     0 &    1 \\
  \ldots is not an infrequent user of Neutrogena shampoo.   & $\circ$       &      0 &     0 &    1 \\
       \textit{}{others}		  		&   & \textit{0}   & \textit{66}  & \textit{85}\\ 
      \bottomrule
   \end{tabular}
   \end{center}
   \caption{Absolute frequency of predicted completions for the hand-written [Hermes] query by three different models. Completions are -- relative to the premises -- entailed ($\star$), redundant ($=$), contradictory ($\dagger$) or independent ($\circ$). }
   \label{table:hermes}
\end{table}


Table~\ref{table:hermes} counts and compares the most frequent completions generated by two {\small TRAIN01} models (762M and 117M) and by the large untrained {\small BASE} model (762M). Exclusively the 762M-model trained on the core schemes reliably predicts the correct conclusion. The large {\small BASE} model rather repeats a premise or even generate a contradiction, whereas the small {\small TRAIN01} model (117M) changes the topic. This is consistent with and illustrates our previous findings. Remarkably, although both the small and the large {\small TRAIN01} models have been fine-tuned on precisely the same arguments, only the large model seems to correctly recognize the logical structure of the [Hermes] argument. Generic language modeling skill, it is suggested, facilitates the successful generalization of learned argument patterns beyond the templates used to create the synthetic training data.    

To further understand transfer learning effects, we next examine whether intermediary pre-training on the artificial argument corpus improves zero-shot performance in other NLP reasoning tasks (i.e., without task-specific fine-tuning).

\paragraph{GLUE AX}

\begin{figure*}[htbp]
    \centering
        \includegraphics[width=.245\textwidth]{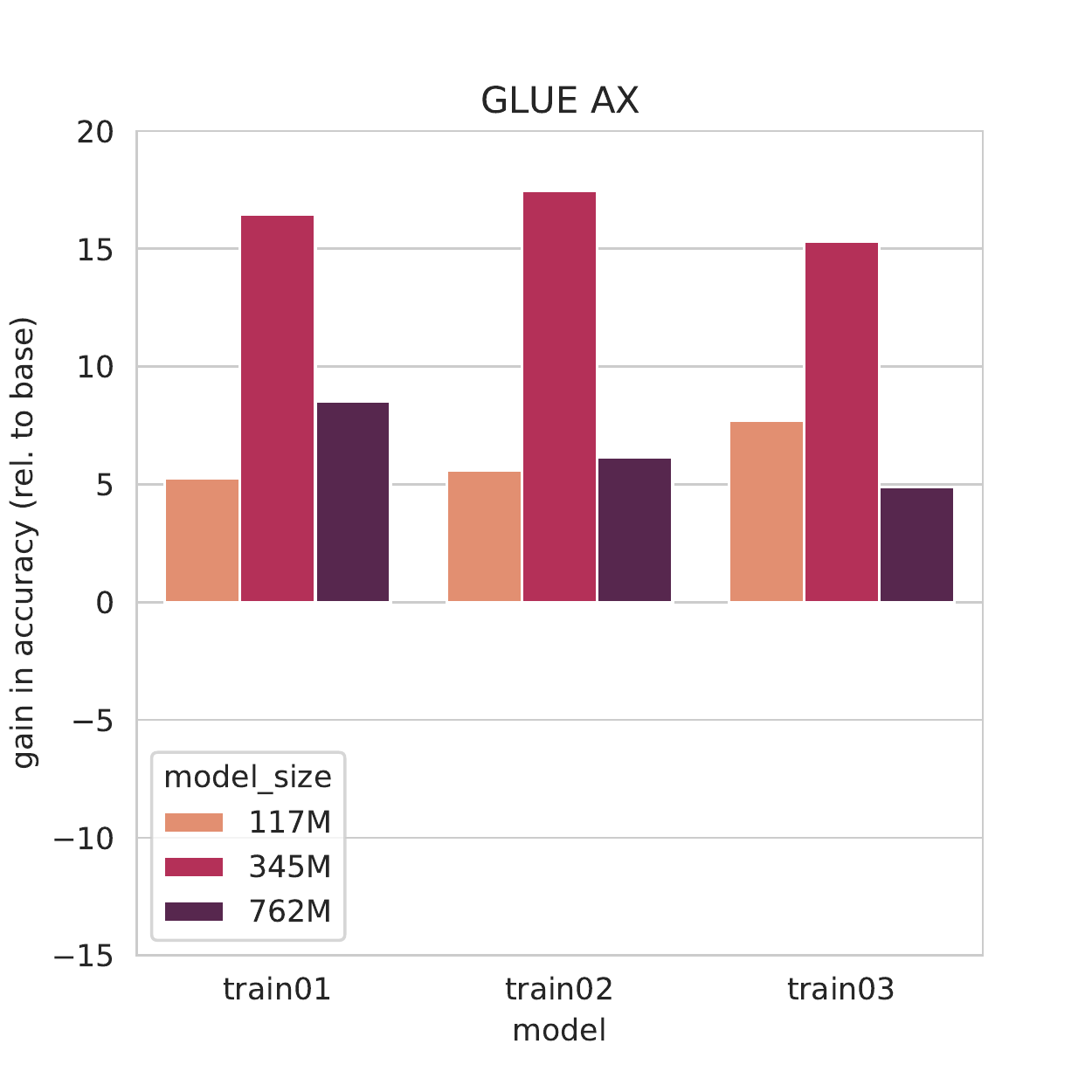}
        \includegraphics[width=.245\textwidth]{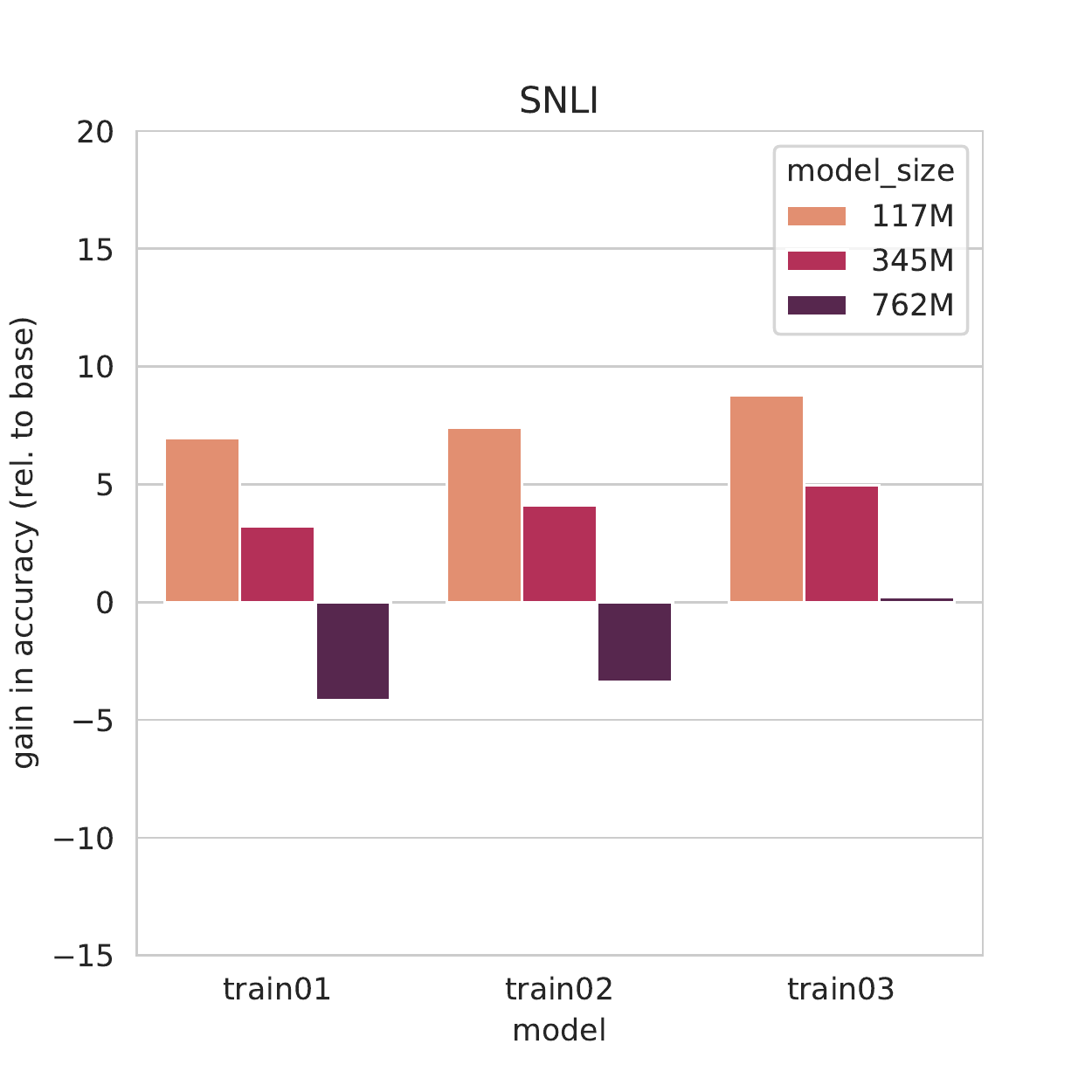}
        \includegraphics[width=.245\textwidth]{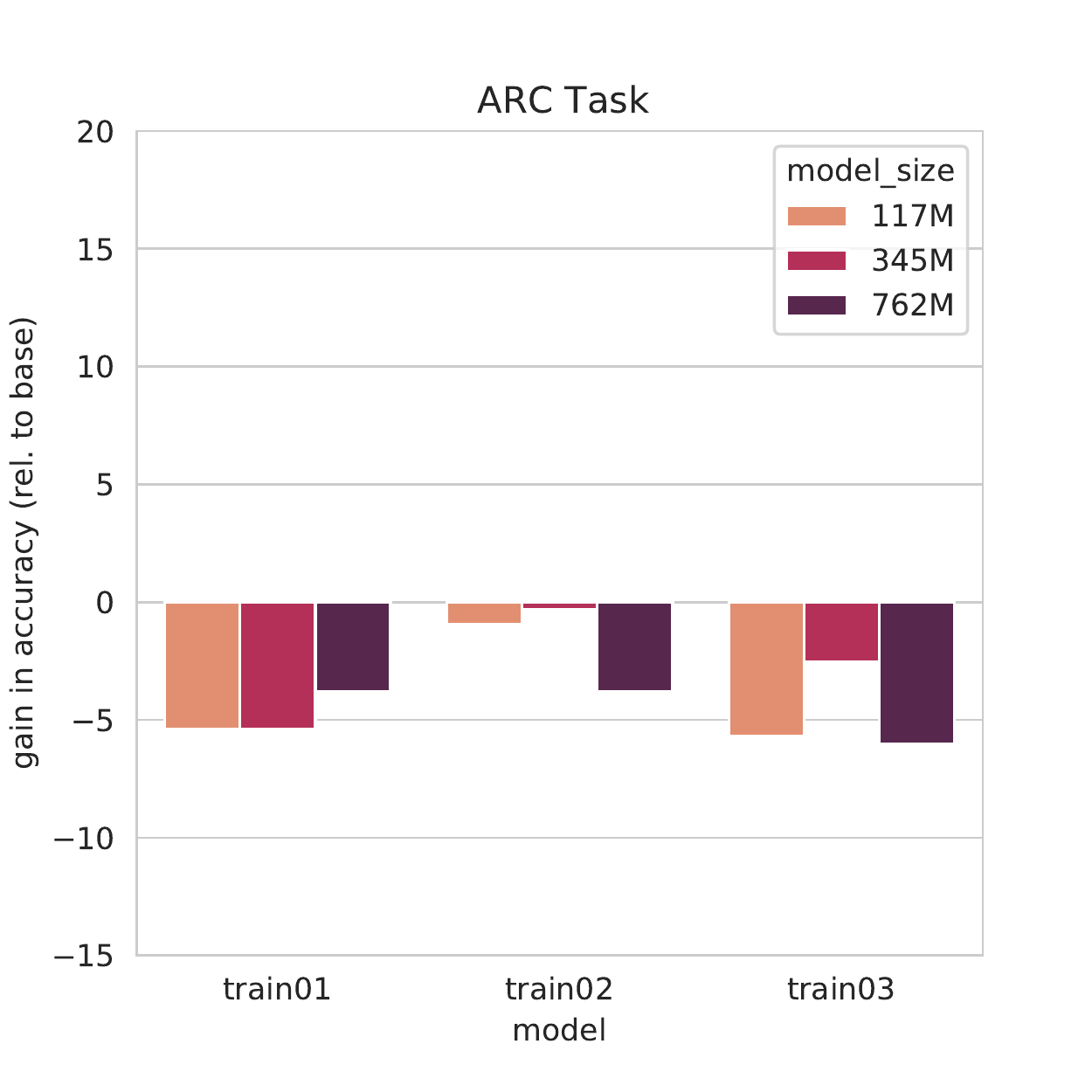}
        \includegraphics[width=.245\textwidth]{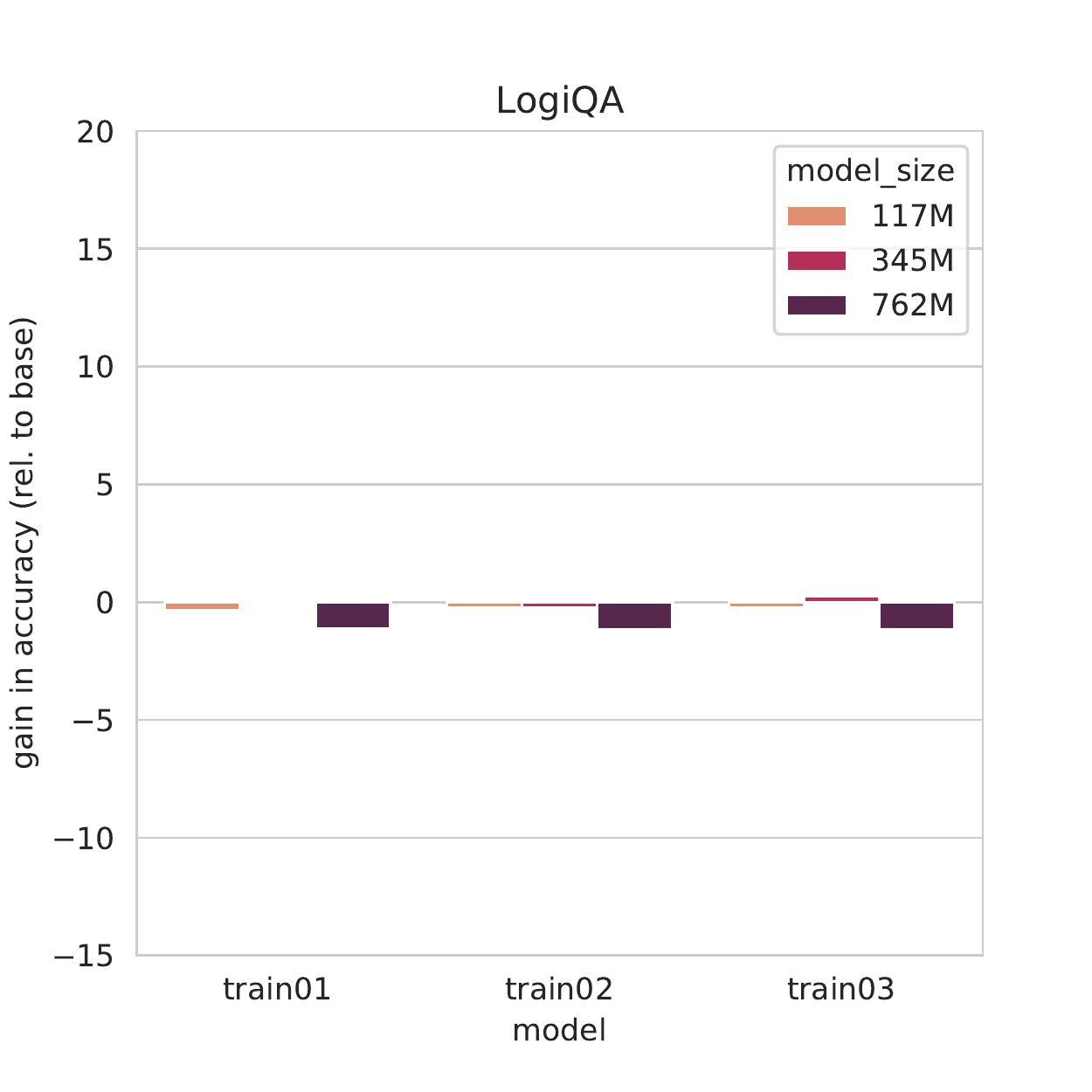}
    \caption{Gains in accuracy due to fine-tuning on the AAC (accuracy {\small TRAIN} model -- accuracy {\small BASE} model) for differently sized models and different NLP benchmark tasks: the GLUE diagnostics data, the SNLI dataset, the argument reasoning comprehension (ARC) benchmark, and the LogiQA dataset.}
    \label{fig:NLP_benchmarks}
\end{figure*}

The GLUE datasets \cite{wang2018glue} represent standard benchmarks for natural language understanding (NLU). We evaluate our models' NLU skill in terms of accuracy on the curated GLUE diagnostics dataset (Figure~\ref{fig:NLP_benchmarks}).

Training on the artificial argument corpus substantially boosts accuracy on the GLUE diagnostics. Accuracy increases by at least 5 and up to 17 percentage points, depending on model size. Remarkably, training on the core scheme alone suffices to bring about these improvements. 

This is a major finding and our clearest evidence so far that training on the AAC involves substantial transfer learning effects.

\paragraph{SNLI}

The SNLI dataset \cite{snli:emnlp2015} is another standard benchmark for NLI. Like the GLUE dataset, it consists in pairs of sentences which entail, contradict, or don't bear on each other. The assessment of our models with respect to SNLI data proceeds in close analogy to the GLUE benchmark.


The results, reported in Figure~\ref{fig:NLP_benchmarks}, are consistent with, albeit less definite than our previous findings for the GLUE benchmark: First and foremost, fine-tuning on all schemes ({\small TRAIN03}) improves the performance by up to 8 percentage points. Training on fewer schemes is slightly less effective. However, it is only the small and medium sized model that profit from fine-tuning on the AAC; the SNLI performance of the 762M parameter model gets rather deteriorated. This might be due to a coincidentally strong performance of the corresponding {\small BASE} model (see Figure~\ref{fig:training_size-NLI-PPL}), or suggest that the large model, unlike the smaller ones, has already learned during pre-training whatever is of relevance for SNLI in the AAC. (Further experiments, preferably involving more model versions, are required to clarify this.)

\paragraph{Argument Reasoning Comprehension Task}

The Argument Reasoning Comprehension (ARC) task \cite{DBLP:conf/naacl/HabernalWGS18a} assesses the ability to identify a missing premise in an informally reconstructed and not necessarily deductively valid argument. It is a multiple-choice task where two alternative sentences are provided, one of which is the missing premise. 

We design and apply specific templates to construct prompts and completions, and calculate relative perplexity as described in Section~\ref{subsec:testing}. 

As shown in Figure~\ref{fig:NLP_benchmarks}, we find no evidence of transfer learning effects with respect to ARC.

\paragraph{LogiQA}

LogiQA \citep{DBLP:conf/ijcai/LiuCLHWZ20} is a collection of nearly 9,000 multiple-choice questions (four alternative answers each) used in critical thinking assessments. These questions span the whole range of critical thinking tasks. 

We design and apply specific templates to construct prompts and completions (one prompt and four completions per question), and use perplexity scores to predict classifications as described above (Section~\ref{subsec:testing}).

As can be seen from Figure~\ref{fig:NLP_benchmarks}, training on the artificial argument corpus has no effect whatsoever on the ability of the models to handle the critical thinking tasks collected in LogiQA.

\section{Conclusion}

This paper has taken a first step towards the creation of a critical thinking curriculum for neural language models. It presents a corpus of deductively valid, artificial arguments, and uses this artificial argument corpus to train and evaluate {\small GPT-2}. The observation of strong transfer learning effects/generalization is its main finding: Training a model on a few central core schemes allows it to accurately complete conclusions of different types of arguments, too. The language models seem to connect and to generalize the core argument schemes in a correct way. Moreover, the models are equally able to apply learned argument patterns beyond the domain they have been trained on, and there is evidence that generic language modeling skill facilitates the successful generalization of learned argument patterns. These findings are consistent with previous work on rule reasoning \cite{Clark2020_TransSoftReas}. They suggest that there exist (learning-wise) fundamental reasoning skills in the sense that generic intermediary pre-training on texts which exemplify these skills leads to spill-over effects and can improve performance on a broad variety of reasoning tasks. The synthetic argumentative texts might be a good starting point for building such a ``critical thinking curriculum for language models.''

Moreover, the trained models have been tested on different reasoning benchmarks. We obtain clear and promising results for the GLUE and SNLI benchmarks. But training on the argument corpus doesn't affect the performance with regard to the semantically more demanding Argument Reasoning Comprehension task or the critical thinking assessment compiled in LogiQA.

Our work suggests different directions for advancing the approach adopted in this paper and further improving the general reasoning skill of neural language models:

\begin{itemize}
  \item The syllogistic argument text corpus might be complemented with corpora of arguments that instantiate \emph{different kinds of correct schemes}, e.g., propositional inference schemes, modal schemes, argument schemes for practical reasoning, complex argument schemes with intermediary conclusions or assumptions for the sake of the argument, etc. (Technically, we provide the infrastructure for doing so, as all this might be achieved through adjusting the argument corpus configuration file.)     
  \item To succeed in NLI tasks, it doesn't suffice to understand `what follows.' In addition, a system needs to be able to explicitly discern contradictions and \emph{non sequiturs} (relations of logical independence). This suggests that the artificial argument corpus might be fruitfully supplemented with corpora of correctly identified aporetic clusters \cite{rescher1987aporetic} as well as corpora containing correctly diagnosed fallacies. 
  \item In addition, the idea of curriculum learning for ML \cite{Bengio:2009} might be given a try. Accordingly, a critical thinking curriculum with basic exemplars of good reasoning would not only be used to fine-tune a pre-trained model, but would be employed as starting point for training a language model from scratch. 
\end{itemize}
  
Natural language templating is a fundamental technique used throughout this paper: both in constructing the artificial argument corpus as well as in transforming the NLP benchmark datasets into text that can be processed by language models. The concrete templates applied have been designed in a trial-and-error process. It is far from clear that these represent optimal choices for effectively eliciting a language model's skills. Still, following \cite{Jiang2020}, it seems of great importance to gain a more systematic understanding of different templating strategies and their effects on metrics based on accuracy and perplexity.     
  
In conclusion, designing a critical thinking curriculum for neural language models seems to be a promising and worthwhile research program to pursue.

\appendix

\section{Appendix: Training Parameters}\label{appendix:hyperparameters}

We train the models on 8 GPUs for 2 epochs with batch size = 2, learning rate = $5\times 10^{-5}$, gradient accumulation steps = 2, and default parameters of the HuggingFace implementation otherwise \citep{wolf2019huggingface}.

\section{Appendix: Performance Metrics for Differently Sized Training Sets}\label{appendix:training_size}

\begin{figure*}[htbp]
    \centering
        \includegraphics[width=.32\textwidth]{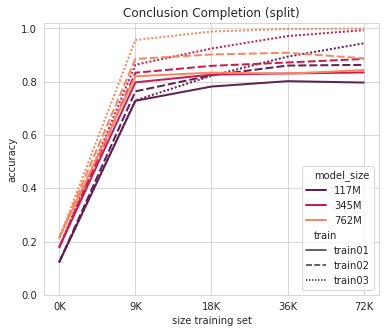}
        \includegraphics[width=.32\textwidth]{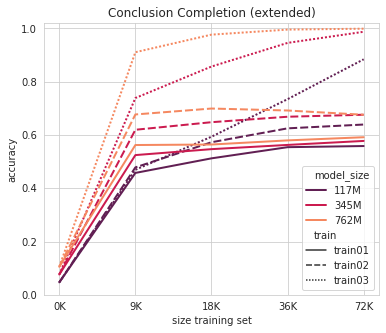}
        \includegraphics[width=.32\textwidth]{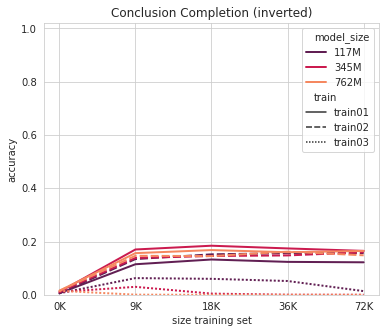}
    \caption{Accuracy on three conclusion completion tasks as a function of training corpus size.}
    \label{fig:training_size-cc}
\end{figure*}

Figure~\ref{fig:training_size-cc} displays accuracy values on conclusion completion tasks for models trained on differently sized datasets. 

\begin{figure*}[htbp]
    \centering
        \includegraphics[width=.32\textwidth]{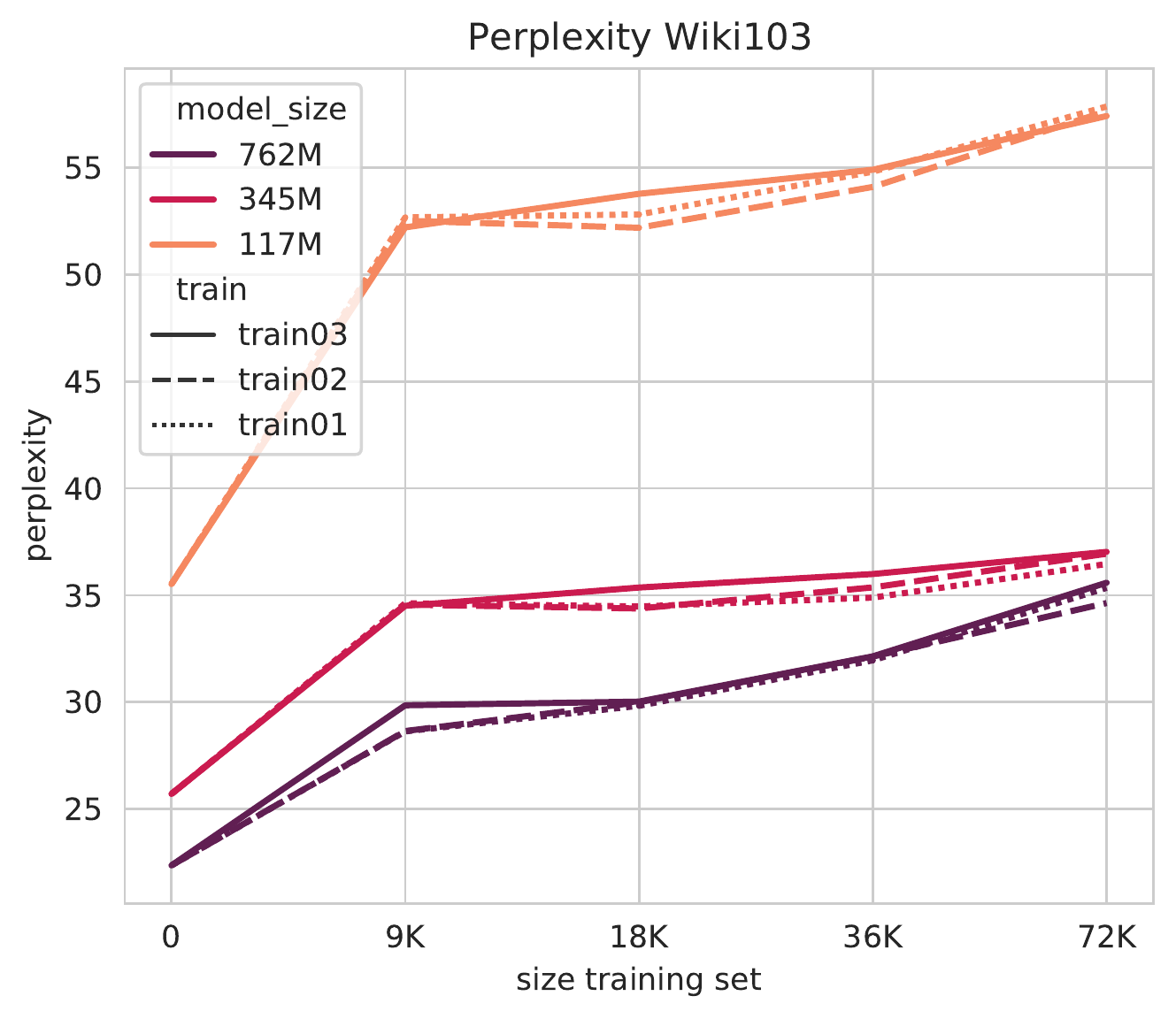}
        \includegraphics[width=.32\textwidth]{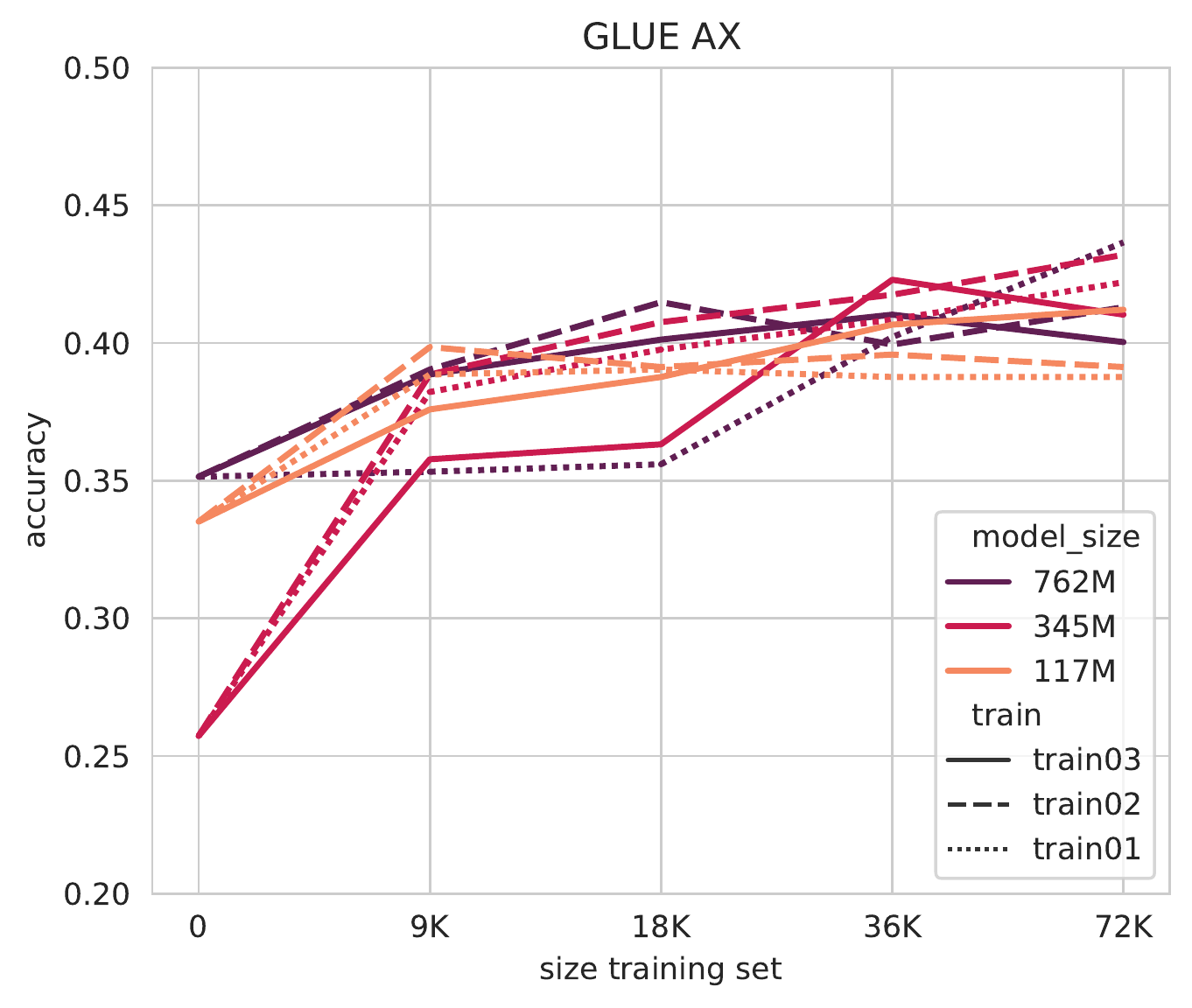}
        \includegraphics[width=.32\textwidth]{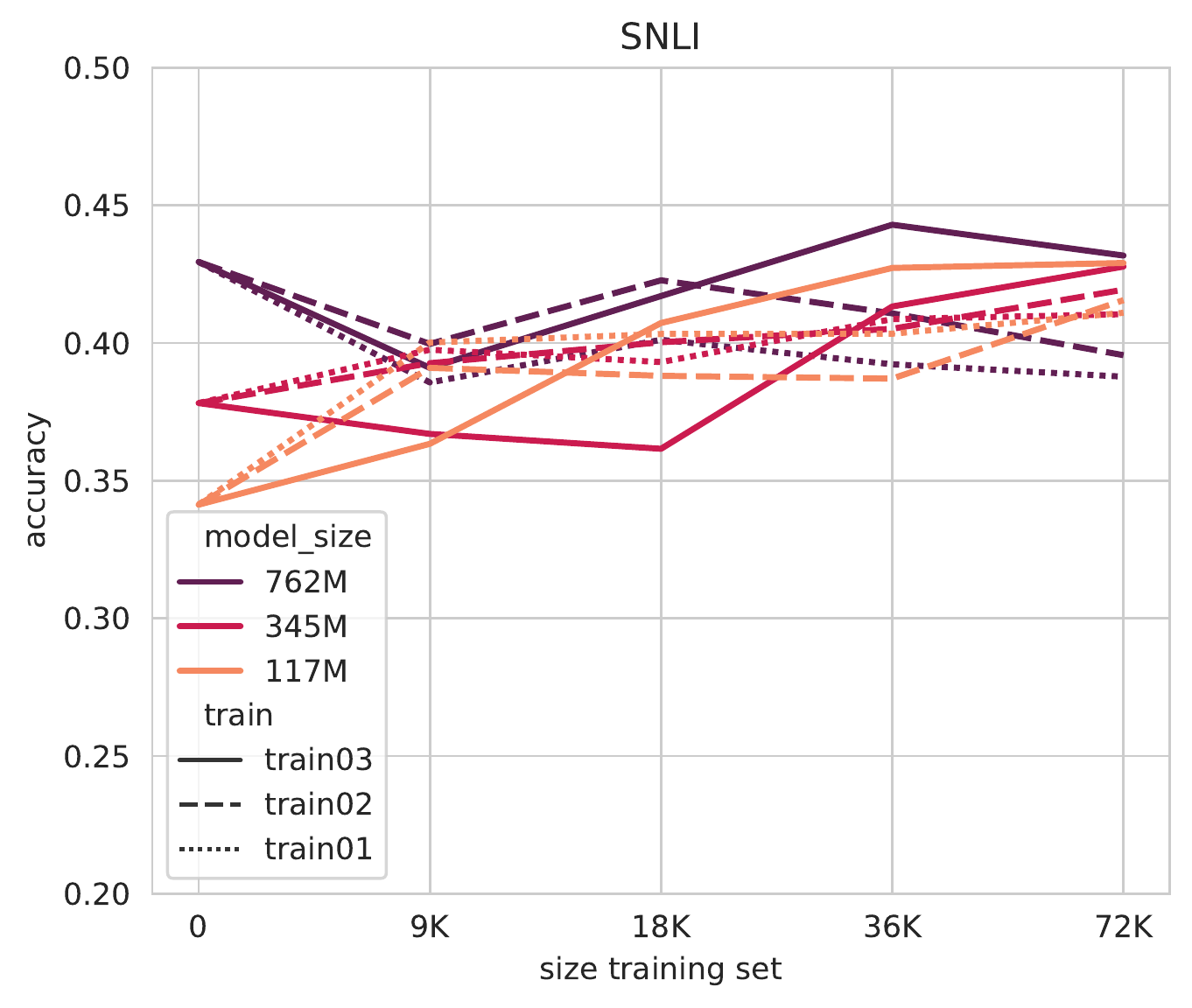}
    \caption{Perplexity and NLI metrics as a function of training corpus size.}
    \label{fig:training_size-NLI-PPL}
\end{figure*}

Figure~\ref{fig:training_size-NLI-PPL} reports perplexity and NLU accuracy metrics for models trained on differently sized datasets.

\bibliography{bib_all.bib}
\bibliographystyle{acl_natbib}

\end{document}